\pdfoutput=1

\documentclass[11pt]{article}

\usepackage{acl}

\usepackage{times}
\usepackage{latexsym}
\usepackage{graphicx}
\usepackage{xcolor}
\usepackage{booktabs}
\usepackage{multirow}
\usepackage{longtable}
\usepackage{url}
\usepackage{hyperref}
\usepackage{caption}
\usepackage{bbm}
\usepackage{subcaption}
\usepackage{amsmath}

\usepackage[T1]{fontenc}

\usepackage[utf8]{inputenc}

\usepackage{microtype}

\newcommand*{\Scale}[2][4]{\scalebox{#1}{$#2$}}%

\title{ConDA: Contrastive Domain Adaptation for AI-generated Text Detection}

\author{Amrita Bhattacharjee ~~ Tharindu Kumarage ~~ Raha Moraffah ~~ Huan Liu \\
        School of Computing and AI \\ Arizona State University \\
        \texttt{\{abhatt43, kskumara, rmoraffa, huanliu\}@asu.edu}}

\begin{document}
\maketitle
\begin{abstract}
Large language models (LLMs) are increasingly being used for generating text in a variety of use cases, including journalistic news articles. Given the potential malicious nature in which these LLMs can be used to generate disinformation at scale, it is important to build effective detectors for such AI-generated text. Given the surge in development of new LLMs, acquiring labeled training data for supervised detectors is a bottleneck. However, there might be plenty of unlabeled text data available, without information on which generator it came from. In this work we tackle this data problem, in detecting AI-generated news text, and frame the problem as an unsupervised domain adaptation task. Here the domains are the different text generators, i.e. LLMs, and we assume we have access to only the labeled source data and unlabeled target data. We develop a \textbf{\underline{Con}}trastive \textbf{\underline{D}}omain \textbf{\underline{A}}daptation framework, called \textbf{ConDA}, that blends standard domain adaptation techniques with the representation power of contrastive learning to learn domain invariant representations that are effective for the final unsupervised detection task. Our experiments demonstrate the effectiveness of our framework, resulting in average performance gains of $31.7\%$ from the best performing baselines, and within $0.8\%$ margin of a fully supervised detector. 
All our code and data is available \href{https://github.com/AmritaBh/ConDA-gen-text-detection}{here}.
\end{abstract}

\section{Introduction}

In recent years there have been significant improvements in the area of large language models that are capable of generating human-like text. Several variants of such language models are designed for specific tasks such as summarization, translation, paraphrasing, etc. Recent advancements in conversational language models such as ChatGPT and GPT-4~\cite{openai2023gpt} have demonstrated how these language models can generate incredibly human-like text, along with serving as an AI assistant for several use cases such as creative writing, explanation of ideas and concepts, code generation and correction, solving mathematical proofs etc.~\cite{bubeck2023sparks}. However, along with improved progress in machine generation of text, there is also a growing concern about how these technologies may be misused and abused by malicious actors. Given how convincing some of these machine-generated texts are, malicious actors may use these models to propagate misinformation/disinformation~\cite{zellers2019defending}, propaganda~\cite{varol2017online}, or even spam/scams. With the accessibility and ease of use of newer language models that have public-facing APIs, the risk of these technologies being used for generating disinformation or misleading information at scale has increased significantly~\cite{de2023chatgpt} and hence has prompted researchers to worry about detection and mitigation strategies~\cite{zhou2023synthetic}. 
\begin{figure}
    \centering
    \includegraphics[width=\columnwidth]{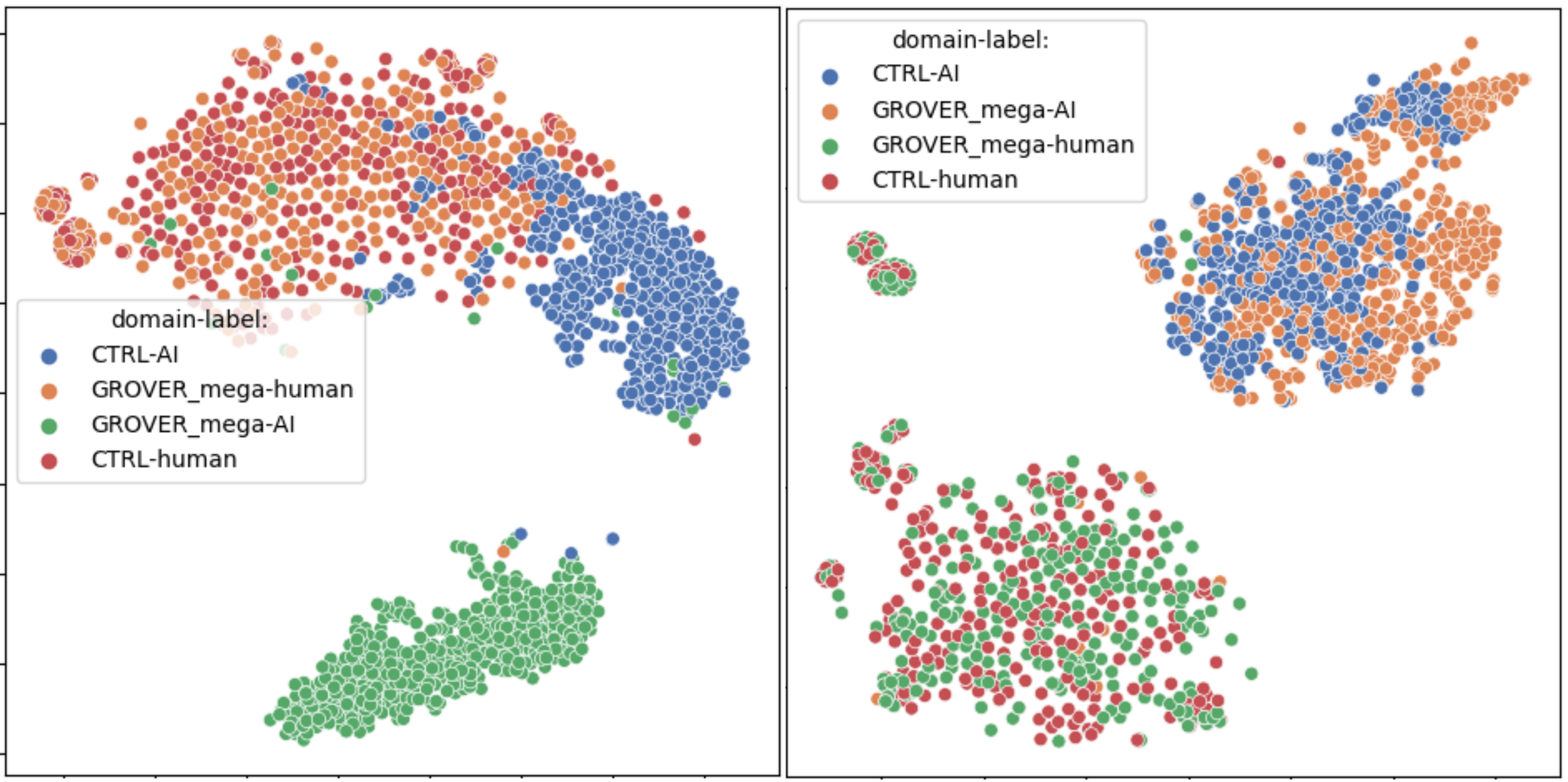}
    \caption{Text embeddings from (\textit{left}) source-only model and (\textit{right}) \textbf{ConDA} model on target domain CTRL with GROVER\_mega as source. Each domain has both `human' and `AI' text. \textbf{ConDA} effectively removes domain-specific features while retaining task-specific features, increasing the separability between `human' and `AI' text, and decreasing the separability between source and target domains.}
    \label{fig:my_label}
\end{figure}
For example, recently, there have been concerns about misleading news websites hosting fully AI-generated news articles\footnote{https://www.newsguardtech.com/special-reports/newsbots-ai-generated-news-websites-proliferating/}. Such unprecedented improvement in language generation capabilities hence naturally necessitates the development of detectors that can accurately and reliably classify such generated text. Motivated by this, we focus on the sub-problem of AI-generated news detection.

A major issue surrounding building a supervised classifier for AI-generated text is the sheer variety of large language models that are available for use. Prior work~\cite{jawahar2020automatic} has demonstrated that detectors built to identify text generated by a particular generator struggle with text from other generators. Furthermore, for newer generators, it might even be impossible to collect and curate labeled training datasets, since access to such models might be limited or even forbidden. Given 
this data problem, in this paper, we consider the situation where we have access to text from a generator but we do not know which generator it came from. However, we do have labeled data from some generators. In this context, we propose a framework for AI-generated text detection that can perform well on target data in the absence of labels. We frame this problem as an unsupervised domain adaptation problem, assuming we have labeled data from a source generator and unlabeled data from (perhaps newer) target generators. 
Our framework also uses a contrastive loss component that acts as a regularizer and helps the model learn invariant features and avoid overfitting to the particular generator it was trained on, hence improving performance on the unknown generator (Figure \ref{fig:my_label}). For news text, our model achieves performance with a $0.8\%$ margin of a fully supervised detector. 
Our main contributions in this paper are:
\begin{enumerate}
    \item We propose a novel AI-generated text detection framework, \textbf{ConDA}, that uses unsupervised domain adaptation and self-supervised contrastive learning to effectively leverage labeled source domain and unlabeled target domain data.
    \item Through extensive evaluations on benchmark human/AI-generated news datasets, spanning a variety of LLMs, we show that \textbf{ConDA} effectively solves the problem of label scarcity, and achieves state-of-the-art performance for unsupervised detection.
    \item Furthermore, we create our own ChatGPT-generated data and via a case study, show the efficacy of our model on text generated using new conversational language models.
\end{enumerate}

\begin{figure*}
    \centering
    \includegraphics[width=\textwidth]{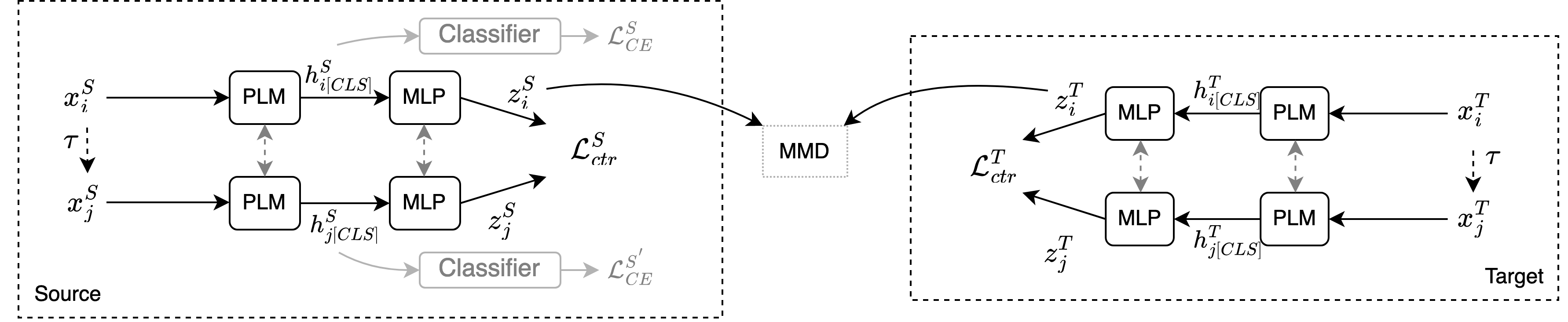}
    \caption{Our ConDA framework. PLM refers to the pre-trained language model (here, RoBERTa); PLM and MLP weights are shared across all four instances.}
    \label{fig:frame}
\end{figure*}

\section{Related Work}


\paragraph{Generated Text Detection} The burgeoning progress in the generation capabilities of large language models has led to a corresponding increase in research and development efforts in the field of detection. Several recent efforts look at methods, varying from simple feature-based classifiers to fine-tuned language model-based detectors, in order to classify whether a piece of input text is human-written or AI-generated ~\cite{ippolito2019automatic,gehrmann2019gltr,mitchell2023detectgpt}, along with methods that specifically focus on AI-generated news~\cite{zellers2019defending,bogaert2022automatic,bhattacharjee2023fighting,kumarage2023stylometric}. A related direction of work is that of authorship attribution (AA). While older AA methods focused on human authors, more recent efforts~\cite{uchendu2020authorship, munir2021through} build models to identify the generator for a particular input text. Recent work also shows how AI-generated text can deceive state-of-the-art AA models~\cite{jones2022you}, thus making the task of detecting such text even more important. 

\paragraph{Contrastive Learning for Text Classification} Following the success of contrastive representation learning in the computer vision domain, several recent works in natural language have used contrastive learning for text classification, often for benefits such as robustness~\cite{zhang2022correct,ghosh2021contrastive,pan2022improved}, generalizability~\cite{tan2020detecting,kim2022generalizable} and also in few-shot scenarios~\cite{jian2022contrastive,zhang2021few,chen2022contrastnet}. Authors in ~\cite{qian2022contrastive,chen2022dual} also use ideas from contrastive learning to leverage label information to learn better representations for the classification task. 

\paragraph{Domain Adaptation for Text Classification}

Domain adaptation (DA) is a paradigm that aims to tackle the distribution shift between training and testing distributions, by learning a discriminative classifier, that is invariant to domain-specific features~\cite{sener2016learning}. Along with labeled source data, DA methods may use either unlabeled target data (unsupervised DA) or a few labeled target samples (semi-supervised DA). In our work, we consider the unsupervised DA setting~\cite{ganin2016domain}. In the domain of language, unsupervised domain adaptation has been used in a variety of tasks~\cite{ramponi2020neural}, such as sentiment classification~\cite{glorot2011domain,trung2022unsupervised}, question answering~\cite{yue2021contrastive}, event detection~\cite{trung2022unsupervised}, sequence tagging or labeling~\cite{han2019unsupervised}, etc.

In this work, we frame the problem of detecting AI-generated news text from multiple generators as an unsupervised domain adaptation task, where the different generators are the different data domains. Our proposed framework combines the representational power of self-supervised contrastive learning and a principled method for unsupervised domain adaptation to solve the AI-generated text detection problem. To the best of our knowledge, we are the first to propose this kind of a formulation for AI-generated text detection, along with a novel framework for this task. In the following section, we describe our framework in detail, along with our training objective. 

\section{Model}

In this work, we consider a setting where we have labeled data from the source generator and only unlabeled samples from the target generator\footnote{We use the terms `LLM' and `generator' interchangeably.}. More formally, the source domain dataset is denoted by $\mathbf{S}=\{(x^S_i,y^S_i)\}_{i=1}^{N^S}$ where $y^S_i \in \{0,1\}$ corresponding to `human-written' or `AI-generated' labels, and $N^S$ is the number of source domain samples. The target domain is denoted by $\mathbf{T}=\{(x^T_i)\}_{i=1}^{N^T}$, where $N^T$ is the number of target domain samples. Note that all domains share the same label space.

\subsection{ConDA Framework}

We show our framework in Figure \ref{fig:frame}. For the detector, we use a pre-trained RoBERTa model (\texttt{roberta-base}) from Huggingface\footnote{https://huggingface.co/roberta-base}, with a classifier head on top of it. As the input, we have two articles: $x^S_i$ from the source and $x^T_i$ from the target. We perform a text transformation $\tau$ on this text whereby we get the transformed samples $x^S_j$ and $x^T_j$. In order to input both the original and the transformed (also referred to as `perturbed' throughout this paper), we use a Siamese network~\cite{bromley1993signature,neculoiu2016learning,reimers2019sentence} where the RoBERTa model weights are shared across the two branches. For the two input texts, we take the hidden layer representation of the \texttt{[CLS]} token: $h_{i[CLS]}^S$ and $h_{j[CLS]}^S$. Following the methodology in ~\cite{chen2020simple}, we pass these embeddings through a projection layer that consists of a multi-layer perceptron (MLP) with one hidden layer and compute a contrastive loss in the lower dimensional projection space. The MLP can be represented as a function $g(\cdot): \mathbbm{R}^{d_h} \mapsto \mathbbm{R}^{d_p}$, where $d_h$ is the size of the hidden layer embedding: $768$ for \texttt{roberta-base}, and we set $d_p$ as $300$, following~\cite{pan2022improved}. For the source domain, we also compute the cross-entropy losses for binary classification of both the original and transformed text. Furthermore, we have a domain discrepancy component between the projected representations of the source and target text. We elaborate on the losses and related design choices in the following section.

\subsection{Training Objective}

\paragraph{Source Classification Loss:} We leverage the availability of the source labels and compute the binary cross-entropy (CE) losses for the original and the perturbed text:
\begin{equation}
\label{eq1}
\begin{split}
    \mathcal{L}_{CE}^{S} & = - \frac{1}{b} \sum_{i=1}^b [y_i \log p(y_i|h^S_{i[CLS]}) + \\
    & (1-y_i) \log(1-p(y_i|h^S_{i[CLS]}))]
\end{split}
\end{equation}


 $\mathcal{L}_{CE}^{S}$ denotes the CE loss for the original text, $b$ denotes the batch size. Similarly, we compute $\mathcal{L}_{CE}^{S^{'}}$ for the perturbed text, and we skip the equation for brevity. Inspired by the training objective in ~\cite{pan2022improved}, we use CE losses for both the original and perturbed samples in the final training objective. The transformation performed on the original text (i.e. synonym replacement in our experiments) preserves the semantics of the text and hence is label-preserving. In such a case we would want a classifier to be able to detect text with such minor, semantic-preserving perturbations as well. Not only is this supposed to improve the robustness of the classifier, but in turn also the generalizability of the detector ~\cite{xu2012robustness}, which is essential for our use-case.

\paragraph{Contrastive Loss:} To learn a better representation of the input text, we use contrastive losses, for both the source and target texts (Figure \ref{fig:frame}). We use a loss similar to the one in ~\cite{chen2020simple}: the only difference is that, instead of computing the loss between two transformed views of the text, we use the transformed text and the original anchor text. For our transformation, we use synonym replacement (more details regarding implementation are in the Appendix). The contrastive loss for the source is denoted by:
\begin{equation}
\label{eq3}
\Scale[0.9]{\mathcal{L}^S_{ctr} = - \sum_{(i,j) \in b} \log \frac{exp(sim(z^S_i,z^S_j)/t)}{\sum_{k=1}^{2|b|} \mathbbm{1}_{[k \neq i]} exp(sim(z^S_i,z^S_k)/t)}}
\end{equation}

 $z^S_i$ and $z^S_j$ denote the projection layer embeddings for the original (anchor) and the transformed text, $t$ is the temperature, $b$ is the current mini-batch, $sim(\cdot,\cdot)$ is a similarity metric which is cosine similarity in our case. Similar to ~\cite{chen2020simple}, we do not sample or mine negatives explicitly, we simply consider the remaining $2(|b|-1)$ samples in the mini-batch $b$ as negatives. We have a similar contrastive loss for the target domain, denoted by $\mathcal{L}^T_{ctr}$, and we skip the equation here for brevity. The objective of these contrastive losses is to bring the positive pairs, i.e. anchor and the transformed sample, closer in the representation space, and well separated from the negative samples. 

Since the performance of contrastive learning depends significantly on the transformation used to generate the positive sample~\cite{tian2020makes}, we take a principled approach to choosing a transformation out of several possible ones~\cite{bhattacharjee2022text}. To choose one transformation for the main experiments, we evaluate a simple detection model (only one domain) over different choices of transformations and choose the one that gives the best performance, and therefore, is the most discriminative. In the input space, we use random swap, random crop, and synonym replacement as the choices. In the latent space, we have paraphrasing and summarization as the choices. Based on detection performance, we finally choose synonym replacement as the transformation that we use throughout the remainder of the paper.


\paragraph{Maximum Mean Discrepancy(MMD):} 
Maximum Mean Discrepancy (MMD)~\cite{gretton2012kernel} is a metric to measure the distance between two distributions, which in our case refers to two different generators. Formally, let $S = \{ x^S_1, x^S_2, ..., x^S_{N^S}\}$ and $T = \{ y^T_1, y^T_2, ..., y^T_{N^T}\}$ be two sets of samples drawn from distribution $\mathcal{S}$ and $\mathcal{T}$, respectively. The MMD distance between the distributions $\mathcal{S}$ and $\mathcal{T}$ is defined as the distance between means of two samples mapped to the Reproducing Kernel Hilbert Space (RKHS)~\cite{steinwart2001influence}. Following past work~\cite{pan2010domain,long2015learning}, we compute the MMD between text embeddings in a lower dimensional space, i.e. between $z^S_i$ and $z^T_i$. Formally,
\begin{equation}
\label{eq4}
\Scale[0.85]{MMD(\mathcal{S}, \mathcal{T}) = \| \frac{1}{N^S} \sum_{i=1}^{N^S} \phi(z^S_i) - \frac{1}{N^T} \sum_{i=1}^{N^T} \phi(z^T_i) \|_\mathcal{H}},
\end{equation}
where $\phi\colon\mathcal{S}\mapsto\mathcal{H}$ and $\mathcal{H}$ represents the RKHS space.

The final training objective for our main framework is:
\begin{equation}
\label{eq5}
\begin{split}
    \mathcal{L} = & \frac{(1 - \lambda_1)}{2} [L^S_{CE} + L^{S^{'}}_{CE}] + \\
                  & \frac{\lambda_1}{2}[L^S_{ctr} + L^T_{ctr}]  + \lambda_2 MMD(S,T)   
\end{split}
\end{equation}

where $\lambda_1$ and $\lambda_2$ are hyper-parameters.

\section{Experimental Settings}

In this section we describe the datasets, baselines and the training details we use for our experiments.

\subsection{Dataset}
\label{sec:data}

Since our task requires news text from multiple generators, we use the publicly available TuringBench\footnote{https://turingbench.ist.psu.edu/} dataset~\cite{uchendu2021turingbench}, which contains human-written and machine-generated news articles from 19 generators, spanning over 10 different language model architectures (including different sizes for some of the generators). For a full list of labels, check Appendix \ref{ap:tb-lab}.
\begin{table}[]
\centering
\resizebox{\columnwidth}{!}{%
\begin{tabular}{@{}ccc@{}}
\toprule
Model        & \# of Parameters & Shorthand used \\ \midrule
CTRL         & 1.5B             & C              \\
FAIR\_wmt19  & 656M             & F19            \\
GPT-2\_xl    & 1.5B             & G2X            \\
GPT-3        & 175B             & G3             \\
GROVER\_mega & 1.5B             & GM             \\
XLM          & 550M             & X              \\ \bottomrule
\end{tabular}%
}
\caption{List of generators we used for our evaluation.}
\label{tab:gens}
\end{table}
Out of the 10 different architectures available in the dataset, we sample a representative set of 6 different generators, in order to evaluate our model (Table \ref{tab:gens}). For most of the architectures, if there were multiple parameter sizes available, we choose the largest one, to make the detection task more challenging for our model. We briefly go over the architectural details of each of the generators used: 

\textbf{CTRL}~\cite{keskar2019ctrl} is a transformer-based language model, that is developed for controllable generation of text based on control codes for style, content, and task-specific generation. The model is pre-trained on a variety of text types, including web-text, news, question-answering datasets, etc. \textbf{FAIR\_wmt19}~\cite{ng2019facebook} is FAIR's model that was developed for the WMT19 news translation task. Texts in TuringBench are from the English version of the FAIR\_wmt19 language model. \textbf{GPT2-XL}~\cite{radford2019language} is the 1.5B size version of GPT-2, which is also a transformer-based language model built upon the architecture of the original GPT model~\cite{radford2018improving}, with further modifications. \textbf{GPT-3}~\cite{brown2020language} is the successor of the GPT-2 model, and is the largest model we use in our evaluation, with a size of 175B parameters. \textbf{GROVER\_mega}~\cite{zellers2019defending} is the largest version of the GROVER model, which is a transformer-based model, similar in architecture to GPT-2, but trained to conditionally generate news articles. \textbf{XLM}~\cite{lample2019cross} is also a transformer-based language model designed for cross-lingual tasks.

Furthermore, given the challenge of detecting text from the more recent conversational language models, we augment the TuringBench dataset with ChatGPT news articles. Following a similar data generation procedure as in ~\cite{uchendu2021turingbench}, we use a subset of around $9,000$ news articles from The Washington Post and CNN (more details in Appendix \ref{ap:tb-human}), and use the headlines to generate articles using ChatGPT. For this paper, we used the OpenAI API with the \texttt{gpt-3.5-turbo} model (version as on March 14, 2023). After experimenting with a few different prompt types, we finally used the following prompt for each news \texttt{headline}: ``Generate a news article with the headline `<\texttt{headline}>'." Finally, we have a balanced dataset of approximately 9k human-written articles, and 9k articles generated using ChatGPT (after accounting for null values and API request errors). For simplicity, we name this dataset \textbf{ChatGPT News} and we use this dataset for a case study on ChatGPT generated news articles, in Section \ref{sec:case-study}.

\subsection{Baselines}
\label{sec:base}

For a fair comparison, we compare our method with baselines that do not require labeled data. We use two open-source AI-generated text detectors, namely GLTR ~\cite{gehrmann2019gltr} and the more recent DetectGPT~\cite{mitchell2023detectgpt}, as our unsupervised baseline models.

\textbf{GLTR} utilizes a proxy language model to calculate the token-wise log probability of the input text. It employs four statistical tests: (i) log probabilities $(\log p(x))$, (ii) average token rank (Rank), (iii) token log-rank (LogRank), and (iv) predictive entropy (Entropy). The first test assumes that a higher average log probability in the input text indicates AI generation. The second and third tests follow a similar assumption, where input texts with lower average rank are more likely to be generated by AI. The last test is based on the hypothesis that AI-generated texts tend to exhibit less diversity and surprises, resulting in low entropy.

\textbf{DetectGPT} also utilizes a proxy language model to calculate the token-wise log probability. However, its decision function is based on comparing the log probability of the original input text with the log probability of a set of $n$ perturbed versions of the input text. These perturbations are generated using the mask-filling language model T5(T5-base)~\cite{raffel2020exploring}. The decision function assumes that if the log probability difference between the input text and the perturbed text is positive with high probability, then the input text is likely to be AI-generated.

In addition to these zero-shot baselines, we include the off-the-shelf \textbf{OpenAI-GPT2 detector} as one of the baselines in our study. The OpenAI-GPT2 detector is a RoBERTa model fine-tuned specifically for detecting GPT2-generated text. It was trained on a GPT-2-output dataset \footnote{https://github.com/openai/gpt-2-output-dataset} comprising 250k documents from the WebText test set ~\cite{radford2019language} as human-written text. Then as the AI text, this dataset contains 250k GPT-2 generated text with a temperature of 1 with no truncation and another 250k samples generated with top-k 40 truncation. Note that for our evaluation, this model may be considered unsupervised for all target domains except GPT-2\_xl. 

\begin{table}[]
\centering
\resizebox{\columnwidth}{!}{%
\begin{tabular}{@{}cccccc@{}}
\toprule
\multirow{2}{*}{Task} &
  \multirow{2}{*}{\begin{tabular}[c]{@{}c@{}}Source\\ Only\end{tabular}} &
  \multirow{2}{*}{ConDA} &
  \multirow{2}{*}{$\Delta$F1} &
  \multirow{2}{*}{\begin{tabular}[c]{@{}c@{}}Source Only\\ Avg.\end{tabular}} &
  \multirow{2}{*}{\begin{tabular}[c]{@{}c@{}}ConDA\\ Avg.\end{tabular}} \\
        &    &    &     &                       &                       \\ \midrule
F19 $\rightarrow$ C   & 93 & 96 & 3   & \multirow{5}{*}{57.2} & \multirow{5}{*}{\textbf{84.6}} \\
G2X $\rightarrow$ C   & 61 & 81 & 20  &                       &                       \\
G3 $\rightarrow$ C    & 61 & 69 & 8   &                       &                       \\
GM $\rightarrow$ C    & 22 & 99 & 77  &                       &                       \\
X $\rightarrow$ C     & 49 & 78 & 29  &                       &                       \\ \midrule
C $\rightarrow$ F19   & 40 & 73 & 33 & \multirow{5}{*}{41.2} & \multirow{5}{*}{\textbf{54.8}} \\
G2X $\rightarrow$ F19 & 83 & 83 & 0   &                       &                       \\
G3 $\rightarrow$ F19  & 62 & 63 & 1   &                       &                       \\
GM $\rightarrow$ F19  & 4  & 27 & 23  &                       &                       \\
X $\rightarrow$ F19   & 17 & 28 & 11  &                       &                       \\ \midrule
C $\rightarrow$ G2X   & 77 & 77 & 0   & \multirow{5}{*}{74.4} & \multirow{5}{*}{\textbf{86.6}} \\
F19 $\rightarrow$ G2X & 90 & 98 & 8   &                       &                       \\
G3 $\rightarrow$ G2X  & 73 & 69 & -4  &                       &                       \\
GM $\rightarrow$ G2X  & 81 & 95 & 14  &                       &                       \\
X $\rightarrow$ G2X   & 51 & 94 & 43  &                       &                       \\ \midrule
C $\rightarrow$ G3    & 60 & 81 & 21  & \multirow{5}{*}{64.8} & \multirow{5}{*}{\textbf{83.2}} \\
F19 $\rightarrow$ G3  & 48 & 89 & 41  &                       &                       \\
G2X $\rightarrow$ G3  & 87 & 82 & -5  &                       &                       \\
GM $\rightarrow$ G3   & 74 & 77 & 3   &                       &                       \\
X $\rightarrow$ G3    & 55 & 87 & 32  &                       &                       \\ \midrule
C $\rightarrow$ GM    & 26 & 95 & 69  & \multirow{5}{*}{37.6} & \multirow{5}{*}{\textbf{73.4}} \\
F19 $\rightarrow$ GM  & 10 & 68 & 58  &                       &                       \\
G2X $\rightarrow$ GM  & 66 & 92 & 26  &                       &                       \\
G3 $\rightarrow$ GM   & 67 & 68 & 1  &                       &                       \\
X $\rightarrow$ GM    & 19 & 44 & 25  &                       &                       \\ \midrule
C $\rightarrow$ X     & 81 & 94 & 13  & \multirow{5}{*}{81}   & \multirow{5}{*}{\textbf{90.2}} \\
F19 $\rightarrow$ X   & 77 & 95 & 18  &                       &                       \\
G2X $\rightarrow$ X   & 87 & 94 & 7   &                       &                       \\
G3 $\rightarrow$ X    & 73 & 69 & -4  &                       &                       \\
GM $\rightarrow$ X    & 87 & 99 & 12  &                       &                       \\ \bottomrule
\end{tabular}%
}
\caption{Performance of \textbf{ConDA} on unlabeled target domains. Source-only model for each task S $\rightarrow$ T refers to zero-shot evaluation of a model trained on S and evaluated on test set of T. $\Delta$F1 is increase (or decrease, in a few cases) in F1 scores of the \textbf{ConDA} model over the source-only model. Avg. scores in \textbf{bold} indicate where ConDA out-performs the source-only model.}
\label{tab:main-da}
\end{table}

\section{Results}

To understand and investigate the effectiveness of our model, we try to answer the following research questions:

- RQ1: Does \textbf{ConDA} perform well on unknown target domains in comparison to a source-only model (Table \ref{tab:main-da}) and a supervised model fine-tuned on the target (Table \ref{tab:comp-sup})?

- RQ2: How well does \textbf{ConDA} perform in comparison to unsupervised-baselines (Table \ref{tab:baselines})?

- RQ3: Are each of the loss components beneficial in training (Table \ref{tab:loss-abla})?

All results are reported as an average over 3 training runs with 3 different random seeds.

\subsection{Performance of ConDA on unlabeled target data}

To evaluate the performance of ConDA on each of the target domains, i.e. generators, we first look at how our model improves over a source-only model. Table \ref{tab:main-da} shows the results for this experiment, grouped by target domain. We report F1 scores for the ConDA framework and a source-only model,  along with scores averaged over sources, for each target. The source-only model is a pre-trained RoBERTa (roberta-base) fine-tuned only on the source domain S. The source-only scores provide an estimate of how well a model trained just on the source transfers to the target domain. 
Although a few of the source-only models have satisfactory performance on the target, using our ConDA framework, we achieve performance gains over the source-only model in almost all tasks (rows with positive $\Delta$F1 values). Particularly interesting are the cases where we use a smaller generator as the source, a larger one as the target, and still get high performance gains: 58 F1 points for FAIR\_wmt19 (656M)$\rightarrow$ GROVER\_mega (1.5B), and 41 F1 points for FAIR\_wmt19 (656M)$\rightarrow$ GPT-3 (175B). This may suggest that, with our ConDA framework, even having unlabeled data from newer and possibly larger generators can improve performance if we use a suitable generator as the source.

\begin{table*}[]
\centering
\aboverulesep=0ex
\belowrulesep=0ex
\renewcommand{\arraystretch}{1.3}
\resizebox{\textwidth}{!}{%
\begin{tabular}{@{}ccccccccccccccccc@{}}
\toprule
\multirow{3}{*}{Target} &
  \multicolumn{2}{c}{\multirow{2}{*}{\begin{tabular}[c]{@{}c@{}}Supervised \\ (Fine-tuned RoBERTa)\end{tabular}}} &
  \multicolumn{14}{c}{ConDA model with Source as} \\ \cmidrule(l){4-17} 
 &
  \multicolumn{2}{c}{} &
  \multicolumn{2}{c|}{C} &
  \multicolumn{2}{c|}{F19} &
  \multicolumn{2}{c|}{G2X} &
  \multicolumn{2}{c|}{G3} &
  \multicolumn{2}{c|}{GM} &
  \multicolumn{2}{c|}{X} &
  \multicolumn{2}{c}{\textit{Average}} \\ \cmidrule(l){2-17} 
 &
  \multicolumn{1}{c}{\small F1} &
  \multicolumn{1}{c|}{\small AUROC} &
  \multicolumn{1}{c}{\small F1} &
  \multicolumn{1}{c|}{\small AUROC} &
  \multicolumn{1}{c}{\small F1} &
  \multicolumn{1}{c|}{\small AUROC} &
  \multicolumn{1}{c}{\small F1} &
  \multicolumn{1}{c|}{\small AUROC} &
  \multicolumn{1}{c}{\small F1} &
  \multicolumn{1}{c|}{\small AUROC} &
  \multicolumn{1}{c}{\small F1} &
  \multicolumn{1}{c|}{\small AUROC} &
  \multicolumn{1}{c}{\small F1} &
  \multicolumn{1}{c|}{\small AUROC} &
  \multicolumn{1}{c}{\small F1} &
  \multicolumn{1}{c}{\small AUROC} \\ \midrule
\multicolumn{1}{c|}{C} &
  98 &
  \multicolumn{1}{c|}{1} &
  \multicolumn{2}{c|}{--} &
  96 &
  \multicolumn{1}{c|}{0.998} &
  81 &
  \multicolumn{1}{c|}{0.949} &
  69 &
  \multicolumn{1}{c|}{0.783} &
  \textbf{99} &
  \multicolumn{1}{c|}{\textbf{1}} &
  78 &
  \multicolumn{1}{c|}{0.991} &
  84.6 &
  0.9442 \\ 
\multicolumn{1}{c|}{F19} &
  98 &
  \multicolumn{1}{c|}{0.999} &
  73 &
  \multicolumn{1}{c|}{0.894} &
  \multicolumn{2}{c|}{--} &
  \textbf{83} &
  \multicolumn{1}{c|}{\textbf{0.966}} &
  63 &
  \multicolumn{1}{c|}{0.607} &
  27 &
  \multicolumn{1}{c|}{0.826} &
  28 &
  \multicolumn{1}{c|}{0.766} &
  54.8 &
  0.8118 \\ 
\multicolumn{1}{c|}{G2X} &
  92 &
  \multicolumn{1}{c|}{0.998} &
  77 &
  \multicolumn{1}{c|}{0.946} &
  \textbf{98} &
  \multicolumn{1}{c|}{\textbf{0.998}} &
  \multicolumn{2}{c|}{--} &
  69 &
  \multicolumn{1}{c|}{0.902} &
  95 &
  \multicolumn{1}{c|}{0.991} &
  94 &
  \multicolumn{1}{c|}{0.991} &
  86.6 &
  0.9656 \\ 
\multicolumn{1}{c|}{G3} &
  72 &
  \multicolumn{1}{c|}{0.988} &
  81 &
  \multicolumn{1}{c|}{0.938} &
  \textbf{89} &
  \multicolumn{1}{c|}{0.975} &
  82 &
  \multicolumn{1}{c|}{0.962} &
  \multicolumn{2}{c|}{--} &
  77 &
  \multicolumn{1}{c|}{\textbf{0.982}} &
  87 &
  \multicolumn{1}{c|}{0.981} &
  83.2 &
  0.9676 \\ 
\multicolumn{1}{c|}{GM} &
  98 &
  \multicolumn{1}{c|}{0.996} &
  \textbf{95} &
  \multicolumn{1}{c|}{\textbf{0.988}} &
  68 &
  \multicolumn{1}{c|}{0.961} &
  92 &
  \multicolumn{1}{c|}{0.984} &
  68 &
  \multicolumn{1}{c|}{0.819} &
  \multicolumn{2}{c|}{--} &
  44 &
  \multicolumn{1}{c|}{0.98} &
  73.4 &
  0.9464 \\
\multicolumn{1}{c|}{X} &
  99 &
  \multicolumn{1}{c|}{1} &
  94 &
  \multicolumn{1}{c|}{0.985} &
  95 &
  \multicolumn{1}{c|}{0.999} &
  94 &
  \multicolumn{1}{c|}{0.988} &
  69 &
  \multicolumn{1}{c|}{0.683} &
  \textbf{99} &
  \multicolumn{1}{c|}{\textbf{0.999}} &
  \multicolumn{2}{c|}{--} &
  90.2 &
  0.9308 \\ \bottomrule
\end{tabular}%
}
\caption{Performance of our ConDA model on each of the target domains, with each of the other domains as source. Numbers in \textbf{bold} are the best performing ConDA models for each target domain, i.e. closest to fully supervised performance.}
\label{tab:comp-sup}
\end{table*}

Next, we compare the performance of our model with a fully-supervised detector trained on the target domain in Table \ref{tab:comp-sup}. 
For ConDA, we show the test performance for all target-source pairs. For the supervised model, we use a pre-trained RoBERTa (roberta-base) fine-tuned on the target data. We then evaluate the model on the test set of the same target domain, and essentially this is our upper bound performance. ConDA achieves test performance comparable to fully-supervised models. In particular, for targets CTRL and XLM, ConDA (with GROVER\_mega as source) achieves upper bound performance. For targets GROVER\_mega and GPT-2\_xl, ConDA performs within 3 and 6 F1 points of the fully-supervised model. 

Interestingly, for target generator GPT-3, all the ConDA models perform better than the fully-supervised performance, with the best F1 (from ConDA with source FAIR\_wmt19) being 27 points higher than the supervised performance. Furthermore, when GPT-3 is used as the source domain, we get mediocre performance for all target domains. We suspect that this might be due to the following reason: The GPT-3 data in TuringBench might be noisy and therefore lack good quality, discriminative signals that can guide the detector. The performance improvement that occurs when ConDA is evaluated on GPT-3 as target, with any other domain as source, is possibly due to the effective transfer of discriminative signals from the labeled source data, hence improving the performance on GPT-3 data even in the absence of labels.

\subsection{Performance compared to unsupervised baselines}

\begin{table*}[]
\centering\scriptsize
\resizebox{\textwidth}{!}{%
\begin{tabular}{@{}ccccclclc@{}}
\toprule
\multirow{3}{*}{Target} &
  \multicolumn{4}{c}{GLTR} &
  \multicolumn{1}{c}{\multirow{3}{*}{DetectGPT}} &
  \multirow{3}{*}{\begin{tabular}[c]{@{}c@{}}OpenAI-GPT2 \\ Detector\end{tabular}} &
  \multicolumn{2}{c}{\multirow{2}{*}{ConDA \textit{(ours)}}} \\ \cmidrule(lr){2-5}
 &
  \multirow{2}{*}{log p(x)} &
  \multirow{2}{*}{Rank} &
  \multirow{2}{*}{LogRank} &
  \multirow{2}{*}{Entropy} &
  \multicolumn{1}{c}{} &
   &
  \multicolumn{2}{c}{} \\ \cmidrule(l){8-9} 
    &       &       &       &       & \multicolumn{1}{c}{} &       & \multicolumn{1}{c}{\textit{Avg.}}    & \textit{Max. \tiny (Source)}   \\ \midrule
C   & 0.951 & 0.849 & \textbf{0.956} & 0.379 & 0.793                & 0.366 & 0.9442                     & \textbf{1.00} \tiny (GM)      \\
F19 & 0.558 & 0.618 & 0.546 & 0.656 & 0.5045               & 0.464 & \textbf{0.8118} & \textbf{0.966} \tiny (G2X)    \\
G2X & 0.485 & 0.508 & 0.48  & 0.631 & 0.529                & 0.48  & \textbf{0.9656}                     & \textbf{0.998} \tiny (F19)    \\
G3  & 0.362 & 0.356 & 0.341 & 0.756 & 0.5485               & 0.73  & \textbf{0.9676}                     & \textbf{0.982} \tiny (GM)     \\
GM  & 0.434 & 0.469 & 0.434 & 0.592 & 0.5415               & 0.659 & \textbf{0.9464}                     & \textbf{0.988} \tiny (C)      \\
X   & 0.473 & 0.762 & 0.442 & 0.696 & 0.7355               & 0.873 & \textbf{0.9308}                     & \textbf{0.999} \tiny (GM,F19) \\ \bottomrule
\end{tabular}%
}
\caption{Performance of ConDA in comparison to unsupervised baselines, as AUROC. For ConDA, we report the average AUROC over all sources (for each target) and also the maximum AUROC (across all sources), along with the corresponding source in parentheses. \textbf{Bold} shows superior performance across each target.}
\label{tab:baselines}
\end{table*}

We compare our ConDA framework with relevant unsupervised baselines and report results in Table \ref{tab:baselines}. 
Out of the four GLTR measures ($\log p(x)$, Rank, Log Rank, and Entropy), the first three fare quite well for detecting CTRL-generated text, but performance on other generators is quite poor. DetectGPT, which is the most recent method we evaluate, performs poorly on almost all generators, with some satisfactory performance on CTRL and XLM.
Surprisingly, the OpenAI GPT-2 Detector performs poorly on the GPT-2\_xl data from TuringBench, although it can be considered supervised for this particular target. Finally, we see ConDA outperforms all the baselines in terms of maximum AUROC, and all but one in terms of average AUROC. 

Interestingly, we see that ConDA models trained with GROVER\_mega as the source perform very well for several target domains. This might be because GROVER~\cite{zellers2019defending} was designed and trained in order to generate news articles. Since our task here is to specifically detect human vs. AI written news articles, training models on data generated using GROVER\_mega is useful and this data possibly has good discriminative signals.

\subsection{Ablation: Effectiveness of loss components}


\begin{table}[]
\centering
\resizebox{\columnwidth}{!}{%
\begin{tabular}{@{}ccccccc@{}}
\toprule
\multirow{3}{*}{Model variant} & \multicolumn{6}{c}{Target (avg. across Sources)}                          \\ \cmidrule(l){2-7} 
                               & \multicolumn{2}{c}{C} & \multicolumn{2}{c}{F19} & \multicolumn{2}{c}{G2X} \\ \cmidrule(l){2-7} 
                               & F1       & AUROC      & F1        & AUROC       & F1         & AUROC      \\ \midrule
ConDA \textbackslash{}CEs      & 60.4     & 0.5268     & 41.6      & 0.4914      & 60.25      & 0.4822     \\
ConDA\textbackslash{}contrast  & 62.6     & 0.898      & 44.2      & 0.687       & 85.4       & 0.9594     \\
ConDA\textbackslash{}MMD       & 69.8     & 0.7826     & 39.8      & 0.6272      & 65         & 0.852      \\
ConDA                          & \textbf{84.6}     & \textbf{0.9442}     & \textbf{54.8}      & \textbf{0.8118}      & \textbf{86.6}       & \textbf{0.9656}     \\ \bottomrule
\end{tabular}%
}
\caption{Comparison of different model variants; \textbf{bold} shows best performance. We randomly chose 3 target domains to show in this table due to space constraints.}
\label{tab:loss-abla}
\end{table}

We evaluate variants of the ConDA model, by removing one component at a time and compare these in Table \ref{tab:loss-abla}. \textbf{ConDA \textbackslash{}CEs} removes the two cross-entropy losses, i.e. no supervision even for the source. \textbf{ConDA \textbackslash{}contrast} removes the contrastive loss components for both source and target. \textbf{ConDA \textbackslash{}MMD} removes the MMD loss between source and target. Hence the only component that makes use of the unlabeled target domain data is the target contrastive loss. Finally, \textbf{ConDA} is the full model. We see that the full model outperforms all the variants, implying that all three types of components are essential for detection performance in this problem setting. Combined with source supervision, the contrastive losses and the MMD objective effectively tie the power of self-supervised learning and unsupervised domain adaptation resulting in superior performance across target domains.

\section{A Case Study on ChatGPT}
\label{sec:case-study}

Given recent concerns surrounding OpenAI's ChatGPT and GPT-4~\cite{openai2023gpt}, it is important to create detectors for text generated by these conversational language models. With the incredible fluency and writing quality these language models possess, not only can such text easily fool humans~\cite{else2023abstracts} but can also be extremely difficult for detectors to identify. Even OpenAI's detector struggles to detect AI-generated text reliably\footnote{https://openai.com/blog/new-ai-classifier-for-indicating-ai-written-text}. Hence in this case study, we are interested in evaluating our ConDA framework on ChatGPT-generated news articles, in an unsupervised manner. Since there is no existing dataset of ChatGPT-generated vs. human-written text or news, we create our own dataset as explained in Section \ref{sec:data}. We assign ChatGPT as the unlabeled target domain and assume that we have labeled data from the 6 other generators (Table \ref{tab:gens}). Therefore we emulate a real-world scenario where labeled data from older generators may be available, but it might be hard to find labeled samples for newer LLMs. We sample 4k articles from our \textbf{ChatGPT News} dataset and evaluate the same 3 unsupervised models as in Section \ref{sec:base} (upper row block in Table \ref{tab:chatgpt}), and our ConDA framework over 6 source generators (lower row block in Table \ref{tab:chatgpt}) on this data. For GLTR, we report the average over the 4 statistical measures. Although we see satisfactory performance across most methods, our ConDA framework with source as FAIR\_wmt19 and GPT2\_xl has the best and the second best performance, respectively. However, we would like the reader to note that such good performance on our ChatGPT News dataset does not imply similar performance on any other type of text generated by ChatGPT (see Section \ref{sec:lim}).  For text embedding visualizations from our ConDA model for this ChatGPT case study, check Appendix \ref{ap:viz}.

\begin{table}[]
\centering
\resizebox{\columnwidth}{!}{%
\begin{tabular}{@{}cccccc@{}}
\toprule
\multicolumn{6}{c}{Baselines}                  \\ \midrule
\multicolumn{2}{c}{GLTR(avg.)} & \multicolumn{2}{c}{DetectGPT} & \multicolumn{2}{c}{OpenAI Detector} \\
\multicolumn{2}{c}{0.72925}    & \multicolumn{2}{c}{0.7735}    & \multicolumn{2}{c}{0.715}                 \\ \midrule
\multicolumn{6}{c}{ConDA model with Source as} \\ \midrule
C      & F19    & G2X   & G3    & GM   & X     \\
0.653  & \textbf{0.877}  & \underline{0.831} & 0.679 & 0.73 & 0.626 \\ \bottomrule
\end{tabular}%
}
\caption{Results on our ChatGPT News dataset using unsupervised baselines (upper row) and ConDA (lower row). Scores are AUROC. \textbf{Bold} shows best and \underline{underline} shows second best performance.}
\label{tab:chatgpt}
\end{table}

\section{Conclusion \& Future Work}

In this work, we address the problem of AI-generated text detection in the absence of labeled target data. We propose a contrastive domain adaptation framework that leverages the power of both unsupervised domain adaptation and self-supervised representation learning, in order to tackle the task of AI-generated text detection. Our experiments focus on news text, and show the effectiveness of the framework, as well as superior performance when compared to unsupervised baselines. We also perform a case study to evaluate our framework on our dataset of ChatGPT-generated news articles and achieve satisfactory performance. Our framework can be easily extended to other forms of text beyond news and our results suggest that such a framework may be effectively used for detection of AI-generated text when labels are unavailable, such as in the case of newly emerging generators. Future work can investigate more challenging variations of this problem, such as domain adaptation across multiple unlabeled target generators, generalization to fully unseen generators, etc., along with exploring other types of text such as scientific articles, medical literature, etc.

\section{Limitations}
\label{sec:lim}

\paragraph{Problem Formulation \& Model:} Despite the impressive performance of our ConDA model, there are several limitations that we go over in this section. First, our model and evaluations only focus on news text and performance may vary widely across other types of text such as creative writing, scientific articles, blog-style articles, etc. Second, our model simply tries to detect whether an input news article is generated by an LLM or not. AI generation does not necessarily imply malice. A dimension that our model does not consider is that of factuality: not all AI-generated news is factually inaccurate, and not all human-written news is factually correct. Incorporating factuality, perhaps in the form of a fact-checking module, could possibly improve the usefulness of our model. Third, our model, along with most other AI-generated text detectors, is not explainable. The discrete input space of natural language also makes it difficult to identify specific features that result in detection. Furthermore, given the black-box nature of LLMs, any detector that uses some LLM as the backbone, trade off explainability for performance gains. 

\paragraph{ChatGPT Case Study:} As we elaborated in Section \ref{sec:data}, we create our own ChatGPT-generated news article dataset, following a procedure similar to ~\cite{uchendu2021turingbench}. However, the data we generated is conditioned on the sample of human-written news articles we randomly selected. We suppose the performance of our model on this ChatGPT data hence is dependent on this sample. The high performance scores for ChatGPT-generated articles could also stem from the inherent structure of news articles; our data is specifically constrained to the style of journalistic news articles. Therefore, good performance on our news article dataset for ChatGPT does not necessarily imply similar performance across text from other areas. For this, more thorough evaluation is needed, which would be an interesting direction for future work.

\section{Ethical Considerations}

We go over some of the ethical considerations surrounding this work and similar directions.

\subsection{Potential to Penalize Benign Use of LLMs}

Recent articles have demonstrated how the newer language models including ChatGPT, GPT-4~\cite{openai2023gpt}, Bing Chat\footnote{https://www.bing.com/new}, etc. can be used to improve productivity, spur creative thinking, help with writing essays or cover letters or even explain concepts and help in homework. As these LLMs become more pervasive, standard use of these as writing or brain-storming assistants may become commonplace. In such a case, we may encounter an increasing amount of text generated by these LLMs online. Such text, if used for benign purposes such as the ones mentioned above, should not be penalized by a detector such as ours. This brings another dimension to this already challenging problem: the issue of intent. Flagging AI-generated content without characterizing the intent behind that could wrongfully penalize users of LLMs. Therefore, the nuances surrounding this need to be considered while using such a detector.

\subsection{Danger of Misuse in High Stakes Areas}

We discuss the issue of model misuse, by taking education as an example. Given the accessibility of ChatGPT and other recent AI-text generators, educators have expressed concerns~\cite{tlili2023if} over students cheating or plagiarising via these new technologies. There are already commercial detectors for AI-generated content such as GPTZero\footnote{https://gptzero.me/} and one from Copyleaks\footnote{https://copyleaks.com/ai-content-detector} that educators may use. However, similar to our model, there is always a margin of error on such detectors. Performing plagiarism checks and subsequently implementing punitive action based solely on such detectors may be detrimental in case of false positives. Legitimate work by a student may be misclassified by these detectors, and potentially impact their career. Eventually, this also diminishes trust in these detectors. Hence, before the widespread use of such AI-generated text detectors, thorough studies on error analysis and reliability need to be performed, along with policy changes to accommodate for the rapidly evolving landscape of AI technologies. 
\section*{Acknowledgements}
This work is supported by the DARPA SemaFor project (HR001120C0123), Office of Naval Research (N00014-
21-1-4002), Army Research Office (W911NF2110030) and Army Research Lab (W911NF2020124). The views, opinions and/or findings expressed are those of the authors.

\bibliography{anthology,custom,references}
\bibliographystyle{acl_natbib}

\appendix

\section{Reproducibility}

\subsection{Training Details \& Hyper-parameters Used} 

We perform all experiments using PyTorch, on a single NVIDIA A100 GPU. We use an Adam optimizer with a learning rate of $2 \times 10^{-5}$. All models are trained for 5 epochs with early stopping, to avoid overfitting. We provide the full list of hyper-parameter values in Table \ref{tab:hyper} to facilitate reproducibility.

For values of $\lambda_1$ and $\lambda_2$ in Equation \ref{eq4}, we use 0.5 and 1.0 respectively, after choosing these values from a small hyper-parameter search. We randomly choose 4 tasks: \{F19 $\rightarrow$ G3, C $\rightarrow$ X, G2X $\rightarrow$ GM, and G2X $\rightarrow$ F19\} and evaluate models with $\lambda_1 = \{0.2, 0.5, 0.8\}$ and $\lambda_2 = \{0.2, 0.5, 1.0\}$. We finally choose $\lambda_1 = 0.5$ and $\lambda_2 = 1.0$ based on these evaluation performances.

\subsection{Synonym Replacement Implementation}

In order to implement the synonym replacement transformation, we perturb 10\% of the words in each sentence in an article by replacing these with their synonyms. Out of these words, we only perform synonym replacement for words that have a \texttt{NOUN}, \texttt{ADVERB}, \texttt{ADJECTIVE} or \texttt{VERB} POS tag. Synonyms are based on WordNet Synsets from the nltk\footnote{https://www.nltk.org/} package. If a word has multiple synonyms, we choose one from that list, uniformly at random.

\section{TuringBench Details}

\begin{table}[]
\centering
\resizebox{\columnwidth}{!}{%
\begin{tabular}{@{}ccc@{}}
\toprule
Hyper-parameter & Description                       & Value    \\ \midrule
$\lambda_1$     & \begin{tabular}[c]{@{}c@{}}Weight for both source \& \\ target contrastive losses in final \\ objective function (Eq. \ref{eq4})\end{tabular} & $0.5$ \\ \midrule
$\lambda_2$     & \begin{tabular}[c]{@{}c@{}}Weight for MMD loss in\\ final objective function (Eq. \ref{eq4})\end{tabular}                                     & $1$   \\ \midrule
$t$    & \begin{tabular}[c]{@{}c@{}}Temperature for \\ contrastive loss  in Eq \ref{eq4}\end{tabular}                                                  & 0.5 \\ \midrule
$lr$             & Learning rate                     & $2\times10^{-5}$ \\ \midrule
$epochs$         & Number of epochs for training     & $5$        \\ \midrule
$max\_seq\_len$    & Maximum input sequence length     & $256$      \\ \midrule
$weight\_decay$ & \begin{tabular}[c]{@{}c@{}}Weight decay for \\ Adam optimizer\end{tabular}                                                              & 0   \\ \midrule
$d_p$  & \begin{tabular}[c]{@{}c@{}}Embedding size of the \\ MLP projection space\end{tabular} & $300$  \\ \midrule
$|b|$     & Batch size for training the model & $16 $      \\ \bottomrule
\end{tabular}%
}
\caption{Hyper-parameter values we used for all our experiments.}
\label{tab:hyper}
\end{table}

\subsection{Labels}
\label{ap:tb-lab}
TuringBench~\cite{uchendu2021turingbench} has 200k samples across 20 labels. These labels include `human' and 19 different generators, which are: \{ Human, GPT-1, GPT-2\_small, GPT-2\_medium, GPT-2\_large, GPT-2\_xl, GPT-2\_PyTorch, GPT-3, GROVER\_base, GROVER\_large, GROVER\_mega, CTRL, XLM, XLNET\_base, XLNET\_large, FAIR\_wmt19, FAIR\_wmt20, TRANSFORMER\_XL, PPLM\_distil, PPLM\_gpt2\}. 

\subsection{Human-written Articles}
\label{ap:tb-human}

Human-written news articles in TuringBench are from The Washington Post, CNN, and a Kaggle dataset with CNN news articles from 2014-2020 and The Washington Post news articles from 2019-2020. More details on the TuringBench data are in ~\cite{uchendu2021turingbench}. For the human-written articles in our \textbf{ChatGPT News} dataset, we use a random sample from the dataset of CNN and Washington Post articles as used in TuringBench.

\section{ChatGPT Visualizations}
\label{ap:viz}

\begin{figure*}[h!] 
\begin{subfigure}{0.32\textwidth}
\includegraphics[width=\linewidth]{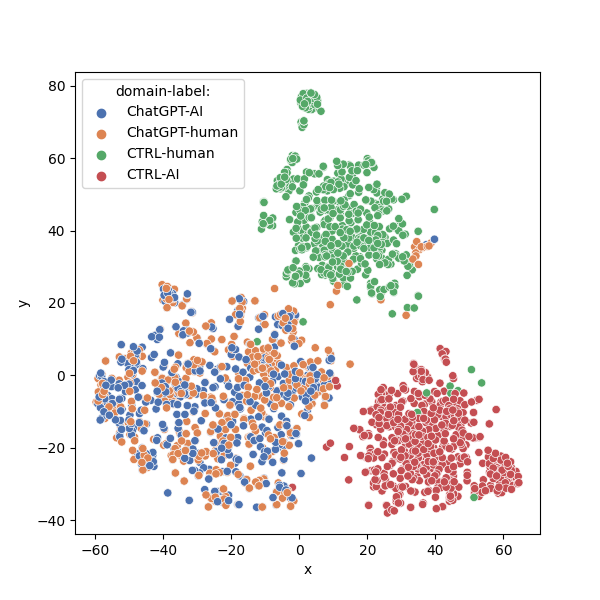}
\caption{CTRL $\rightarrow$ ChatGPT} \label{fig:c}
\end{subfigure}\hspace*{\fill}
\begin{subfigure}{0.32\textwidth}
\includegraphics[width=\linewidth]{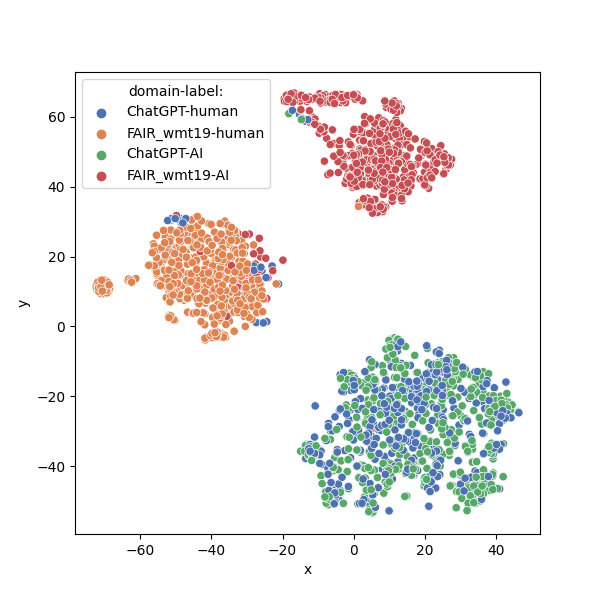}
\caption{FAIR\_wmt19 $\rightarrow$ ChatGPT} \label{fig:b}
\end{subfigure}\hspace*{\fill}
\begin{subfigure}{0.32\textwidth}
\includegraphics[width=\linewidth]{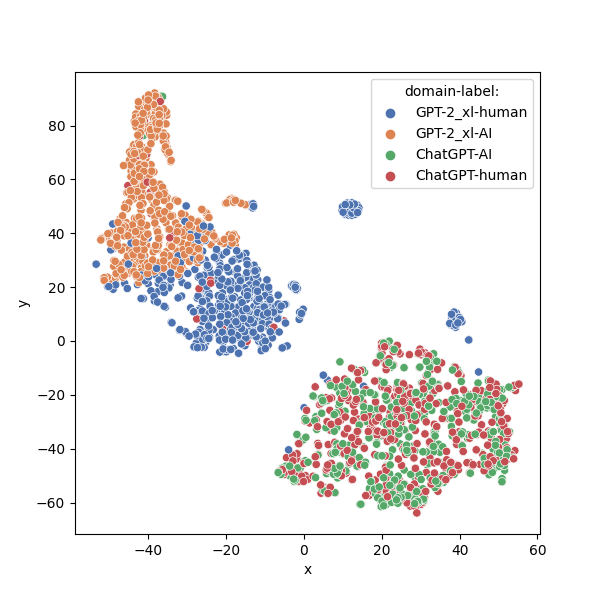}
\caption{GPT-2\_xl $\rightarrow$ ChatGPT} \label{fig:a}
\end{subfigure}\hspace*{\fill}

\medskip
\begin{subfigure}{0.32\textwidth}
\includegraphics[width=\linewidth]{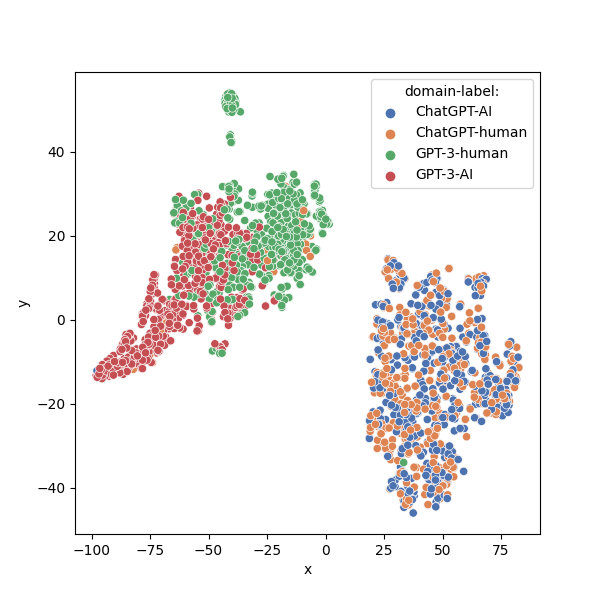}
\caption{GPT-3 $\rightarrow$ ChatGPT} \label{fig:a}
\end{subfigure}\hspace*{\fill}
\begin{subfigure}{0.32\textwidth}
\includegraphics[width=\linewidth]{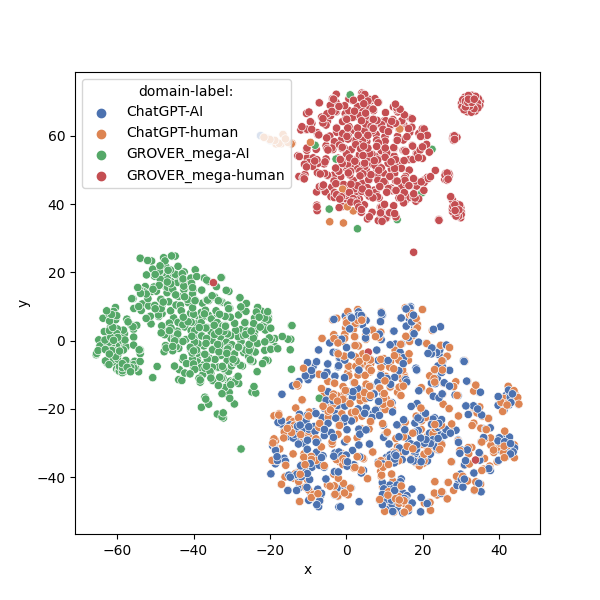}
\caption{GROVER\_mega $\rightarrow$ ChatGPT} \label{fig:b}
\end{subfigure}\hspace*{\fill}
\begin{subfigure}{0.32\textwidth}
\includegraphics[width=\linewidth]{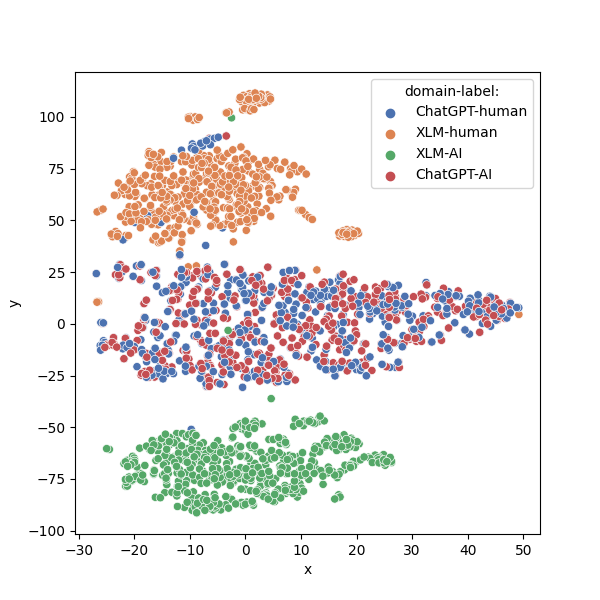}
\caption{XLM $\rightarrow$ ChatGPT} \label{fig:a}
\end{subfigure}
\caption{t-SNE plots showing text representations from our ConDA model, for each of the \textit{S} $\rightarrow$ ChatGPT tasks, where \textit{S} $\in$ \{CTRL, FAIR\_wmt19, GPT-2\_xl, GPT-3, GROVER\_mega, XLM\}, corresponding to plots (\textbf{a}-\textbf{f}), respectively.}
\label{fig:tsne-all}
\end{figure*}

Here, we visually explore embeddings from the ConDA model for instances in Table \ref{tab:chatgpt}, in order to understand the issues surrounding the detection of ChatGPT-generated news articles. Figure \ref{fig:tsne-all} shows the embeddings from all 6 ConDA models. In all the plots, we see that the human-written and ChatGPT-generated news articles in our ChatGPT News dataset are very closely clustered together, and are not separable. Therefore, even though our model achieves substantially high AUROC scores, there are possibly many false positives and/or false negatives, thus providing an intuition that better feature selection methods might be necessary here. 


    
\end{document}